\pdfoutput=1
\documentclass[10pt,twocolumn,letterpaper]{article}

\usepackage{iccv}
\usepackage{times}
\usepackage{epsfig}
\usepackage{graphicx}
\usepackage{amsmath}
\usepackage{amssymb}
\usepackage{subfig}
\usepackage{multirow}
\usepackage{multicol}
\usepackage{booktabs}
\usepackage{pifont}
\newcommand{\cmark}{\ding{51}}%
\newcommand{\xmark}{\ding{55}}%
\usepackage{mathtools}
\usepackage{balance}
\DeclarePairedDelimiterX{\infdivx}[2]{(}{)}{%
  #1\;\delimsize\|\;#2%
}
\newcommand{\infdiv}{KLD\infdivx}

\usepackage[breaklinks=true,bookmarks=false]{hyperref}

\iccvfinalcopy 

\def\iccvPaperID{****} 

\ificcvfinal\fi

\begin{document}

\title{Learning to Diversify for Single Domain Generalization}

\author{Zijian Wang\quad  Yadan Luo \quad  Ruihong Qiu \quad  Zi Huang\quad Mahsa Baktashmotlagh \\
University of Queensland \\
\{zijian.wang, y.luo, r.qiu, m.baktashmotlagh\}@uq.edu.au,  huang@itee.uq.edu.au \\}

\maketitle
\ificcvfinal\thispagestyle{empty}\fi

\begin{abstract}
Domain generalization (DG) aims to generalize a model trained on multiple source (\textit{i.e.}, training) domains to a distributionally different target (\textit{i.e.}, test) domain. In contrast to the conventional DG that strictly requires the availability of multiple source domains, this paper considers a more realistic yet challenging scenario, namely Single Domain Generalization (Single-DG), where only one source domain is available for training. In this scenario, the limited diversity may jeopardize the model generalization on unseen target domains. To tackle this problem, we propose a style-complement module to enhance the generalization power of the model by synthesizing images from diverse distributions that are complementary to the source ones. More specifically, we adopt a tractable upper bound of mutual information (MI) between the generated and source samples and perform a two-step optimization iteratively: (1) by minimizing the MI upper bound approximation for each sample pair, the generated images are forced to be diversified from the source samples; (2) subsequently, we maximize the MI between the samples from the same semantic category, which assists the network to learn discriminative features from diverse-styled images. Extensive experiments on three benchmark datasets demonstrate the superiority of our approach, which surpasses the state-of-the-art single-DG methods by up to $25.14\%$. The code will be publicly available at \textcolor{blue}{\href{https://github.com/BUserName/Learning_to_diversify}{https://github.com/BUserName/Learning\_to\_diversify}}
\end{abstract}

\section{Introduction}
The remarkable success of modern machine learning algorithms is built on the assumption that the source (\textit{i.e.}, training) and target (\textit{i.e.}, test) samples are drawn from similar distributions. In practice, this assumption is commonly violated by various factors, such as the changes of illuminations, object appearance, or background, which are known as the \textit{domain shift} problem \cite{discrepancy1, discrepancybendavid}. Due to the discrepancy across domains, the performance of a model trained on the source domain can be significantly degraded when applied to the target domain.

To tackle this problem, extensive research has been carried out mainly on domain adaptation and domain generalization. Domain adaptation aims to transfer the knowledge from the labeled source domain(s) to an unlabeled target domain \cite{DAsurvey, DIP, DAN, DANN}, while domain generalization attempts to generalize a model to an unseen target domain by learning from multiple source domains \cite{DICA, ADA, MASF, ADA}. Compared with domain adaptation, domain generalization (DG) is considered as a more challenging task as the target samples are not exposed in the training phase. Regarding different strategies for transferring knowledge from source domains to the unseen target domain, existing DG techniques can be subsumed under two broad categories, \textit{i.e.}, \textit{alignment-based} \cite{MetaReg,MetaVIB,CCSA} and \textit{augmentation-based} \cite{JIGSAW,RSC,HEX,L2A, randconv}. Technically, alignment-based approaches aim to reach the consensus from multiple source domains and learn domain-invariant latent representations for the target domain. Augmentation-based approaches learn to augment the source images with different transformations or generate the pseudo-novel samples for each source domain. 

In general, the paradigm of DG relies on the availability of multiple source domains. However, it is more plausible to consider a more realistic scenario namely single domain generalization (Single-DG), where \textit{only one} source domain is at hand during training. Despite DG has been extensively studied, single-DG remains under-explored. It is nontrivial for prior DG methods to handle this new setting, as the samples gathered from multiple sources and the domain identifiers are no longer accessible. Without the domain information to rely on, neither alignment-based nor augmentation-based models could well identify the domain-invariant features or transformations that are robust to unseen target domain shifts. Very recently, a few works \cite{MADA, MEADA} are proposed that learn to add adversarial noises on the source images in order to train robust classifiers against unforeseen data shifts. While contributing positively to address the model vulnerability, the manipulated images are indistinguishable from the original source images, which are not sufficiently diverse to cover the real target distributions.

\begin{figure*}[!t]
\centering
    \includegraphics[width=0.995\linewidth]{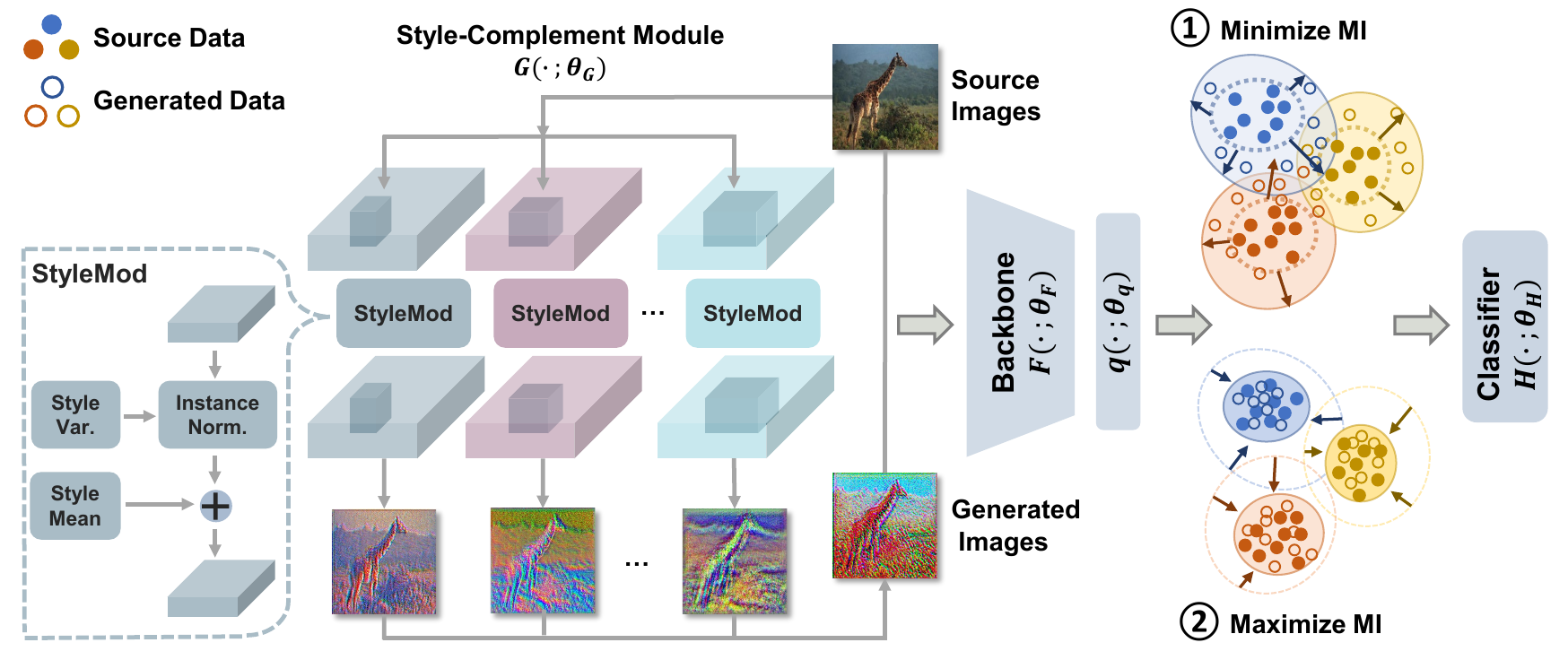}
    \caption{The overall framework of the proposed Learning-to-diversify (L2D). L2D alternatively trains the style-complement module $G(\cdot;\theta_G)$ and the task model $F(\cdot;\theta_F)$, $q(\cdot;\theta_q)$, and $H(\cdot;\theta_H)$. Specifically, (1) the upper bound of mutual information (MI) is minimized between the source and generated images, and (2)MI among samples belonging to the same category is maximized. It enhances the generalization power of the task model in an adversarial min-max manner. }
\label{fig:flowchart}
\end{figure*}

To address the issue of insufficient diversity, in this paper, we propose a novel approach of Learning-to-diversify (L2D), which aims to improve the model generalization capacity by alternating diverse sample generation and discriminative style-invariant representation learning as illustrated in Figure \ref{fig:flowchart}. Specifically, we design a style-complement module, which learns to synthesize samples with unseen styles, which are out of original distributions. Different from previous augmentation approaches that quantify diversity with the Euclidean distance in the image space, we diversify the generated samples in the latent feature space. By explicitly posing a greater challenge on the trained classifier, the model resilience against the target shift is enhanced. We gradually enlarge the shift between the distributions of the generated samples and the source samples at the training time and perform the two-step optimization iteratively. By minimizing the tractable upper bound of mutual information for each sample pair, the generated images are forced to diversify from the source samples in subspace. Furthermore, to obtain the style-invariant features, we maximize the mutual information between the images belonging to the same semantic category. Consequently, the style-complement module and the task model compete in a min-max game, which improves the generalization ability of the task model by iteratively generating out-of-domain images and optimizing the style-invariant latent space. Note that, under this objective, the images generated by the proposed style-complement module will not only being diversified from the source ones, but also can be considered as challenging samples to the task model.

The main contributions of our work are summarized as follows.
\begin{itemize}
\item We propose a style-complement module that addresses the single domain generalization by learning to generate diverse images.

\item A min-max mutual information optimization strategy is designed to gradually enlarge the distribution shift between the generated and the source images, while simultaneously bringing the samples from the same semantic category close in the latent space.

\item To validate the effectiveness of the proposed method, extensive experiments are conducted on three benchmark datasets, including digits recognition, corrupted CIFAR-10, and PACS. We further show the effectiveness of our approach in the standard DG setting under the leave-one-domain-out protocol. The results clearly demonstrate that our method surpasses the state-of-the-art DG and Single-DG methods on all datasets. 
\end{itemize}

\section{Related Work}
\noindent\textbf{Domain Shift.} Most of the existing machine learning methods suffer from performance degradation when the source (\textit{i.e.} training) domain and the target (\textit{i.e.} test) domain follow different distributions. The difference between the distributions is termed as the domain shift \cite{discrepancy1, discrepancy2}. In computer vision applications, such a shift may be brought by but not limited to environment and style changes. To this problem, domain adaptation (DA) approaches have been proposed to minimize the domain shift across the domains by matching the marginal \cite{DAN, DANN, DIP} or conditional distributions \cite{CDAN, PGL} of the source and target domains. Domain adaptation has been widely studied in various settings, such as semi-supervised \cite{semiDA, semiDA2, PMGN} and unsupervised scenarios \cite{DAN, JAN}, which leverage the partially labelled or unlabelled target domain in the training process. More recently, few-shot DA \cite{fewshotDA} is proposed, where only a few labelled target samples together with source samples are available in the training phase.

\noindent\textbf{Domain Generalization.} The most significant difference between domain adaptation (DA) and domain generalization (DG) is that DG does not require access to the target domain in the training phase. Existing DG methods can be rough classified into two categories of learning the domain invariant representation and data augmentation. The key idea behind the former category is to reduce the discrepancy between representations of multiple source domains. Muandet \textit{et al.} \cite{DICA} first proposed a kernel-based method to obtain domain invariant features. \cite{MTAE} learns the latent invariant representation by jointly considering the domain reconstruction task. \cite{CCSA} further introduces a contrastive semantic alignment loss, which encourages the intra-class similarity and inter-class difference. Li \textit{et al.} proposed to reduce the gap across the domains by minimizing Maximum Mean Discrepancy (MMD) under adversarial autoencoder framework. \cite{HEX} aims to learn a domain agnostic representation by enforcing an orthogonal constrain on textural information of images and the corresponding latent representations. Recently, meta-learning procedure has been studied to solve DG problem \cite{MASF, MLDG, MetaVIB}. Li \textit{et al.} put forward a gradient-based model agnostic meta-learning algorithm for DG. Dou \textit{et al.} \cite{MASF} exploits the episodic training scheme to enforce the global and local alignment. \cite{MetaVIB} incorporates variational information bottleneck with meta-learning to narrow the domain gap between the source domains.

The other category is related to data augmentation. This line of work generally aims to generate out-of-domain samples, which are then used to train the network along with the source samples to improve the generalization ability. For instance, \cite{ADA} exploits the adversarial training scheme to generate `hard' samples for the classifier. Shankar \textit{et al.} \cite{CROSSGRAD} proposed to augment the source samples along the direction of the domain change. \cite{L2A} exploits a conditional generative adversarial network (GAN) to synthesize data from pseudo-novel domain. \cite{JIGSAW} is considered as another type of augmentation, which exploits an auxiliary self-supervision training signal from solving a jigsaw puzzle to improve the generalization ability of the classifier.

This paper focuses on a more challenging yet realistic setting, namely single domain generalization \cite{MADA, MEADA}. In Single-DG, the network is trained on a single source domain, while it is evaluated on multiple unseen domains. Gradient-based image augmentation is an effective strategy for Single-DG. \cite{MADA} improves the ADA by encouraging semantic consistency between the augmented and source images in the latent space via an auxiliary Wasserstein autoencoder. \cite{MEADA} considers entropy maximization in the adversarial training framework to generate challenging perturbations of the source samples. In all the aforementioned approaches for Single-DG, the visual differences between the source and generated images are mostly depicted in the color and texture of the augmented samples. Different from existing Single-DG methods, our method aims to generate diverse samples with novel style/texture/appearance having a larger shift from the source distribution, and thus can be considered as complementary to the source data distribution.
\section{Methodology}

Given a source domain $\mathcal{S} = \{x_i, y_i\}_{i=1}^N$ of $N$ samples, the goal of single domain generalization is to learn a model that can generalize to many unseen target domains $\mathcal{T}$. With no prior knowledge on the target domain, we propose a style-complement module $G(\cdot; \theta_G): x \rightarrow x^+$ to augment the source domain by synthesizing the $x^+$ that shares the same semantic information with the source image $x$, albeit with different styles. As shown in Figure \ref{fig:flowchart}, we firstly apply the feature extractor $F(\cdot; \theta_F)$ to transform images $x$ and $x^+$ to latent vectors $z$ and $z^+$. To diversify the generated features from the source samples, the MI upper bound approximation for each pair $z$ and $z^+$ is minimized to learn $G$; subsequently, we freeze $G$ and maximize MI between the $z$ and $z^+$ from the same semantic category, which assists the task network $F$ and classifier $H(\cdot; \theta_H)$ parameterized by $\theta_H$ to learn discriminative features from diverse-styled images.

\noindent\textbf{Style-Complement Module.} The style-complement module $G(\cdot; \theta_G)$ consists of $K$ transformations, each of which is comprised of a convolutional layer, a style learning layer, and a transposed convolutional layer. By applying the convolution operations, the source images are projected from the original distribution to a novel distribution with arbitrary style shift. We further enhance the diversity of the generated samples by creating the style shift at the pixel level. Specifically, we add learnable parameters $\theta_{G,k} = \{\mu_k, \sigma_k\}$ for each of the $k$ transformations, where the learnable parameters $\mu_k, \sigma_k \in \mathbb{R}^{h*w*c}$ are the mean shift and variance shift. More concretely, we have: 
\begin{equation}
    T(f_{i,k}; \theta_{G,k}) = \sigma_k * \frac{f_{i,k}-\mu}{\sigma} + \mu_k,
\end{equation}

\noindent where $f_{i,k} \in \mathbb{R}^{h*w*c}$ is the output of the convolutional operation applied on $x_i$ in the $k$-th transformation, with $h, w$ and $c$ representing the height, width, and channel, respectively. $\mu$ and $\sigma$ correspond to the mean and covariance of the $f_{i,k}$. The transformed feature map $f_{i,k}^\prime = T(f_{i,k}; \theta_{G,k})$ are then applied with the transposed convolutional operation to reconstruct to the original image dimension of $x$. The final outcome $x_i^+$ of the style-complement module is the linear combination of the augmented images $x_{i,k}^+$ obtained from $k$-th transformation:
\begin{equation}
\begin{split} 
    &x_i^+ = \frac{1}{\sum_{k=1}^K(w_k)}\sum_{k=1}^K(w_k \sigma(x_{i,k}^+)), \\
    &w_k \sim Normal(0,1)
\end{split}
\end{equation}
where $w_k$ is a scalar sampled from a normal distribution and weights the contribution of augmented image $x_{i,k}$ from the transformation $k$ to the output augmented image $x_i^+$. We apply  the activation function $\sigma(\cdot)$, \textit{e.g. tanh}, to scale $x_i^+$.

\noindent\textbf{Synthesizing Novel Styles.} The objective of the style-complement module is to generalize from the source domain distribution to the out-of-domain distribution. To increase the diversity of the created styles, the correlation between the generated images and the source images should be minimized. Mutual information (MI) $I(z;z^+)$ serves as a measure of quantifying the correlation of $z$ and $z^+$, which is defined as:
\begin{equation}
    I(z;z^+) = \mathbb{E}_{p(z,z^+)}\big[\log\frac{p(z^+|z)}{p(z^+)}\big]
\end{equation}

We minimize the mutual information (MI) between the source and generated images in the latent feature space $\mathbb{Z}$, achieved by passing images through $F(\cdot; \theta_F)$. The upper bound of MI defined in \cite{CLUB} is:
\begin{equation}
    I(z; z^+) \le \mathbb{E}_{p(z, z^+)}[\log p(z^+ | z)] - \mathbb{E}_{p(z)p(z^+)}[\log p(z^+ | z)],
\end{equation}
where $z$ and $z^+$ are the respective latent vectors of the source image $x$ and the generated image $x^+$. 

Since the conditional distribution $p(z^+ | z)$ is intractable, the upper bound of $I(z;z^+)$ cannot be directly minimized. Therefore, we adopt a variational distribution $q(z^+|z)$, that employs a neural network parameterized by $\theta_q$ to approximate the upper bound $\hat{I}(z;z^+)$ of mutual information:
\vspace{-1ex}
\begin{equation} \label{eq:Min_MI}
    \hat{I}(z;z^+) = \frac{1}{N}\sum_{i=1}^{N}[\log q_{\theta}(z_i^+|z_i) - \frac{1}{N}\sum_{j=1}^{N}\log q_{\theta}(z_j^+|z_i)]
\end{equation}
By minimizing Eq. (\ref{eq:Min_MI}), the learnable mean/variance shift parameters in our model are trained to complement the style of the source domain. Although $\hat{I}(z;z^+)$ is no longer an upper bound for MI as the conditional distribution $p(z^+|z)$ is substituted with a variational approximation $q_\theta(z^+|z)$, $\hat{I}(z;z^+)$ can be a reliable upper bound estimator if the difference between the two distributions is small. Specifically, we estimate the the difference $\Delta$ between $\hat{I}(z;z^+)$ and the upper bound of $I(z;z^+)$ by using the Kullback–Leibler divergence (KLD):
\begin{equation} \label{delta}
\begin{split}
        \Delta &= \infdiv{p(z^+, z)}{q_\theta(z^+,z)}\\
    &= \mathbb{E}_{p(z,z^+)} [\log(p(z^+|z)p(z)) - \log(q_\theta(z^+|z)p(z))]  \\
    &= \mathbb{E}_{p(z,z^+)} [\log p(z^+|z)] - \mathbb{E}_{p(z,z^+)} [\log q_\theta(z^+|z)].
\end{split}
\end{equation}
The above equation shows that the difference $\Delta$ is affected by two terms. Since the first term of Eq. \ref{delta} is not related to $\theta_q$, we minimize the negative log-likelihod between $z_i$ and $z_i^+$ instead of directly minimizing $\Delta$:
\begin{equation}\label{eq:likilhood}
    L_{likeli}= - \frac{1}{N}\sum_{i=1}^{N}\log q_\theta(z^+_i|z_i).
\end{equation}

\noindent\textbf{Semantic Consistency.} While the style-complement module can generate images with diverse styles, it may introduce noise or generate images with distorted semantic information from original source images (\textit{e.g.}, when the variance shift $\sigma_k$ equals to 0, the generated images will become meaningless). Therefore, it is important to limit the conditional distribution shift from the source distribution to the out-of-domain distribution, thereby avoiding generating semantically unrelated images. To achieve this, we minimize the class-conditional maximum mean discrepancy (MMD) in the latent space as follows,
\begin{equation}\label{eq:const}
    L_{const} = \frac{1}{C}\sum_{m=1}^C(\| \frac{1}{n_s^m}\sum_{i=1}^{n_s^m}\phi(z_i^m) - \frac{1}{n_t^m}\sum_{i=1}^{n_t^m}\phi(z^{m+}_j) \|^2),
\end{equation}
\noindent where $z_i^m$ and $z^{m+}_i$ are the $i$-th latent vector of the source and augmented sample of the class $m$, respectively. $n_s^m$ and $n_t^m$ are the total number of original and augmented sample of class $m$. $\phi(\cdot)$ represents the kernel function. Conditional MMD mitigates the potential semantic information distortion by constraining the distribution shift of samples that belong to the same class. 

\noindent\textbf{MI Maximization.} We aim to obtain a generalizable and robust model by playing a min-max game between the style-complement module $G(\cdot; \theta_G)$ and the task model $F(\cdot; \theta_F)$. While the style-complement module aims to generate diverse images that have minimum information on the source images, the task model can cluster images with same semantic label in the embedding space. The lower bound on MI between two varaibles proposed in \cite{infonce} is:
\begin{equation}
    I(z;z^+) \geq \mathbb{E}\big[\frac{1}{N}\sum_{i=1}^N \log \frac{e^{f(z,z^+)}}{\sum_{j=1}^N e^{f(z_i, z^+_j)}}\big],
\end{equation}
where $f(\cdot,\cdot)$ is a critic function.

However, directly maximizing the lower bound MI of the generated and source images without leveraging the semantic label may falsely reduce the shared information of same class samples. To alleviate this problem, we employ the supervised contrastive loss \cite{SUPCON} to increase the mutual information among samples from the same class, defined as:
\begin{equation}\label{eq:max_MI}
\begin{split}
     &L_{supcon} = -\sum_{i=0}^N\frac{1}{|P(i)|} \sum_{p\in P(i)} \log \frac{e^{(z_i\cdot z_p / \tau)}}{\sum_{a\in A(i)} e^{(z_i\cdot z_a / \tau)}}\\
     &P(i) = \{p\in A(i): y_p = y_i\},
\end{split}
\end{equation}
where $A(i)$ is the set of the source and generated latent representations $z, z^+$ of the same class. $\tau$ is the temperature coefficient.

To further enhance the semantic consistency, we minimize the cross-entropy loss on both the source images $X$ and the generated images $X^+$:
\begin{equation}\label{eq:task}
    L_{task} = -\frac{1}{2N}[ \sum_{i=0}^{N} y_i\log(\hat{y_i}) + \sum_{j=0}^{N} y_j^+\log(\hat{y}_j^+)], 
\end{equation}
where $\hat{y}$ and $\hat{y}^+$ are the prediction of the source and generated images, respectively.


\noindent\textbf{Objective Function.} We adopt a two-step training, in which we optimize the style-complement module $G(\cdot;\theta_G)$ and a task model, including $F(\cdot;\theta_F)$, $q(\cdot; \theta_q)$ and $H(\cdot; \theta_H)$ in an iterative manner. Specifically, we train the task module $F$ with both the source images $X$ and the generated images $X^+$ with the weighted combination of Eq. \eqref{eq:Min_MI}, \eqref{eq:likilhood}, and  \eqref{eq:task} as follows:
\begin{equation}
    \min_{\theta_F, \theta_q, \theta_H} L = L_{task} + \alpha_1 L_{supcon} + \alpha_2 L_{likeli}. 
\end{equation}
Notably, $\alpha_1$ and $\alpha_2$ are the hyper-parameters for balancing the losses. To optimize $G$, we consider solving Eq. \eqref{eq:Min_MI} and \eqref{eq:const} jointly:
\begin{equation}
    \min_{\theta_G} L = \hat{I}(z;z^+) + \beta L_{const},
\end{equation}
with $\beta$ being the balancing weight between the mutual information upper bound estimation $\hat{I}(z;z^+)$ and semantic consistent loss $L_{const}$.

\noindent\textbf{Implementation Notes:} To capture multi-scale information from images, we apply different transformations (\textit{i.e.,} kernel sizes) in the convolution and transposed convolution layers. Moreover, to avoid potential distortion of the semantic information, we fix the number of output channels to the number of input color channels (\textit{i.e.,} 3 output channels for RGB images). We re-initialize the weights of the convolutional layer and transposed convolutional layer by sampling from uniform distribution of $(-\frac{1}{\sqrt{size(kernel)}}, \frac{1}{\sqrt{size(kernel)}})$ in each iteration.

\section{Experiments}
\begin{table}[t]
\begin{center}
\caption{Single domain generalization accuracy (\%) comparison on digits dataset. Models are trained on MNIST and evaluated on the rest of the digits datasets. Best performances are highlighted in bold.} 
\label{tab: Digits}
\resizebox{0.495\textwidth}{!}{
\begin{tabular}{l|cccc|c}
\toprule
& SVHN & MNIST-M & SYN & USPS & Avg.\\
\hline
ERM & 27.83 & 52.72 & 39.65 & 76.94 & 49.29\\
CCSA & 25.89 & 49.29 & 37.31 & 83.72 & 49.05\\
d-SNE & 26.22 & 50.98 & 37.83 & \textbf{93.16} & 52.05\\
JiGen & 33.80 & 57.80 & 43.79 & 77.15 & 53.14\\
ADA & 35.51 & 60.41 & 45.32 & 77.26 & 54.62\\
M-ADA & 42.55 & 67.94 & 48.95 & 78.53 & 59.49\\
ME-ADA & 42.56 & 63.27 & 50.39 & 81.04 & 59.32\\
\hline

Ours & \textbf{62.86} & \textbf{87.30} & \textbf{63.72} & 83.97 & \textbf{74.46}\\
\bottomrule

\end{tabular}}
\end{center}
\end{table}

\noindent \textbf{Datasets}. To evaluate the effectiveness of the proposed method, we conduct experiments over three single-dg benchmark datasets. \textbf{Digits} consists of 5 different datasets, which are MNIST\cite{MNIST}, SVHN\cite{SVHN}, MNIST-M\cite{SYN}, SYN\cite{SYN} and USPS\cite{USPS}.  Each dataset is considered as a unique domain which may be different from the rest of the domains in font style, background, and stroke color. \textbf{PACS} \cite{pacs} is a recent proposed DG benchmark dataset that has four domains, including photo, art painting, cartoon, and sketch. Each domain contains $224 \times 224$ images belonging to seven categories, and there are 9,991 images in total. Compared with the digits dataset, PACS is considered as a more challenging dataset due to the large style shift among domains. For fair comparison, we follow the official split of the train, validation, and test. \textbf{Corrupted CIFAR-10} \cite{cifar-c, cifar} contains $32 \times 32$ tiny RGB images from CIFAR-10 that are corrupted by different types of noises. There are 15 corruptions from 4 main categories, including weather, blur, noise, and digital. Each of the corruptions has 5 levels serverities and `5' indicates the severest corruption. 


\subsection{Comparisons on Digits}

\noindent\textbf{Experiment Setup}. Following \cite{ADA, MADA, MEADA}, we select 10,000 images from MNIST as the source domain and test the generalization performance of models on the other four digits datasets. We resize all the images to $32\times32$ and duplicate their channels to convert all the grayscale images to RGB. We employ the LeNet \cite{MNIST} as the backbone and set the batch size as 32. We use SGD to optimize both the style-complement module and the task model. 


\noindent\textbf{Results}. We compare the proposed method with three recent state-of-the-arts single domain generalization methods \cite{ADA, MADA, MEADA}, two multi-domain generalization methods \cite{CCSA, dsne}. Table \ref{tab: Digits} shows that our model achieves the highest average accuracy compared to the other baselines. Specifically, we observe clear improvements of 20.3\%, 29.36\%, 13.33\% and 14.9\% on SVHN, MNIST-M, SYN, and overall accuracy, respectively. Previous Single-DG methods directly generate the auxiliary training samples by applying the adversarial perturbation. Compared to the adversarial gradient-based methods, L2D creates larger domain shifts between the generated images and the source images by incorporating the style-complement module. The substantial increase of the accuracy over the baselines demonstrates the importance of generating diverse-styled images on improving the generalization power of a model. Moreover, we observe that single-DG methods generally achieve better performance in this task, reflecting the dependency of previous DG methods on multiple source domains to learn a generalizable model. Our method achieves the second best performance on USPS. We infer this might be related to the fact the USPS and the source domain MNIST share very similar stroke styles. In such case, the diverse generated images might not benefit the generalizibility of the model as much as the task with a larger domain shift.

\begin{table}[t]
\begin{center}
\caption{Single domain generalization accuracy (\%). Models are trained on CIFAR-10 dataset and evaluated on CIFAR-10-C dataset with corruption severity level 5. We report the average accuracy over 4 main categories of corruption: Weather, Blur, Noise, and Digital. Best performances are highlighted in bold. * indicates our implementation.} 
\label{tab: cifar}
\resizebox{0.47\textwidth}{!}{
\begin{tabular}{l|cccc|c}
\toprule
& Weather & Blur & Noise & Digits& Avg. \\
\hline
ERM & 67.28 & 56.73 & 30.02 & 62.30 & 54.08 \\

CCSA & 67.66 & 57.81 & 28.73 & 61.96 & 54.04 \\

d-SNE & 67.90 & 56.59 & 33.97 & 61.83 & 55.07\\
\hline
ADA$^*$ & 72.67 & 67.04 & 39.97 & 66.62 & 61.58\\

M-ADA & 75.54 & 63.76 & 54.21 & 65.10 & 64.65\\

ME-ADA$^*$ & 74.44 & \textbf{71.37} & 66.47 & 70.83 & 70.77\\

\hline
Ours & \textbf{75.98} & 69.16 & \textbf{73.29} & \textbf{72.02} & \textbf{72.61}\\
\bottomrule
\end{tabular}}
\end{center}\vspace{-3ex}
\end{table}

%


\begin{figure*}[!t]
	\centering
	\subfloat[]
	{\includegraphics[width=0.295\textwidth]{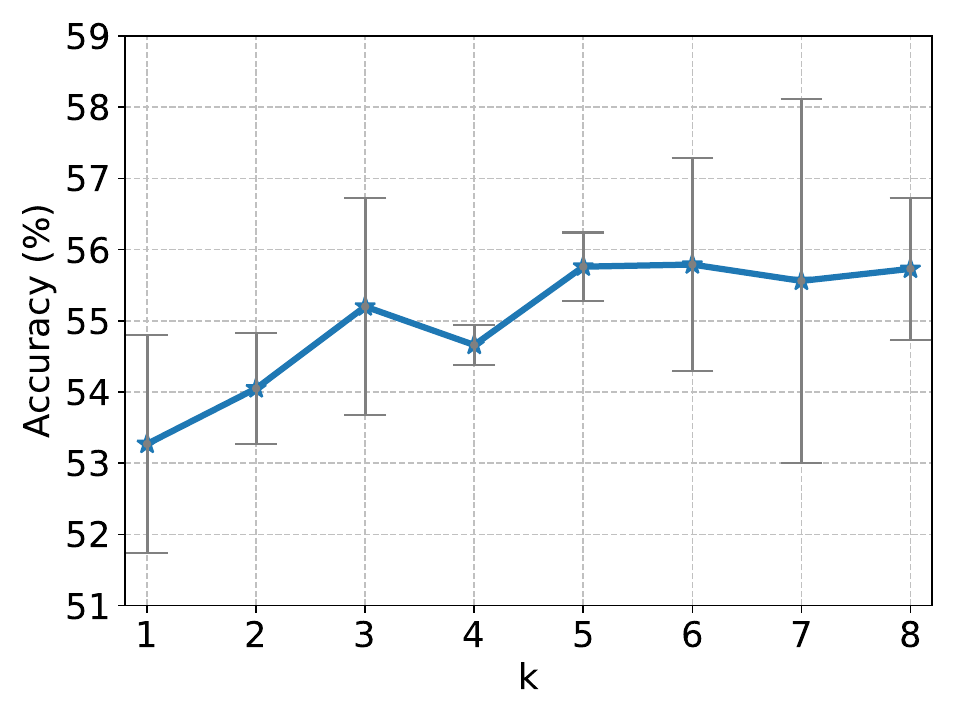}}
	\subfloat[]
	{\includegraphics[width=0.295\textwidth]{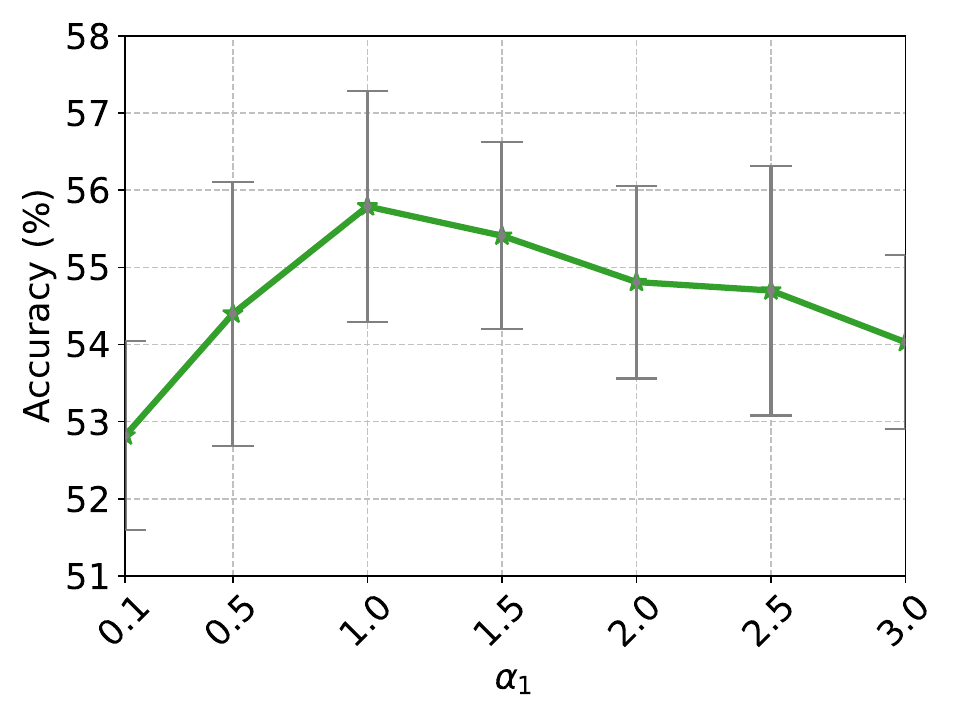}}
	\subfloat[]
	{\includegraphics[width=0.295\textwidth]{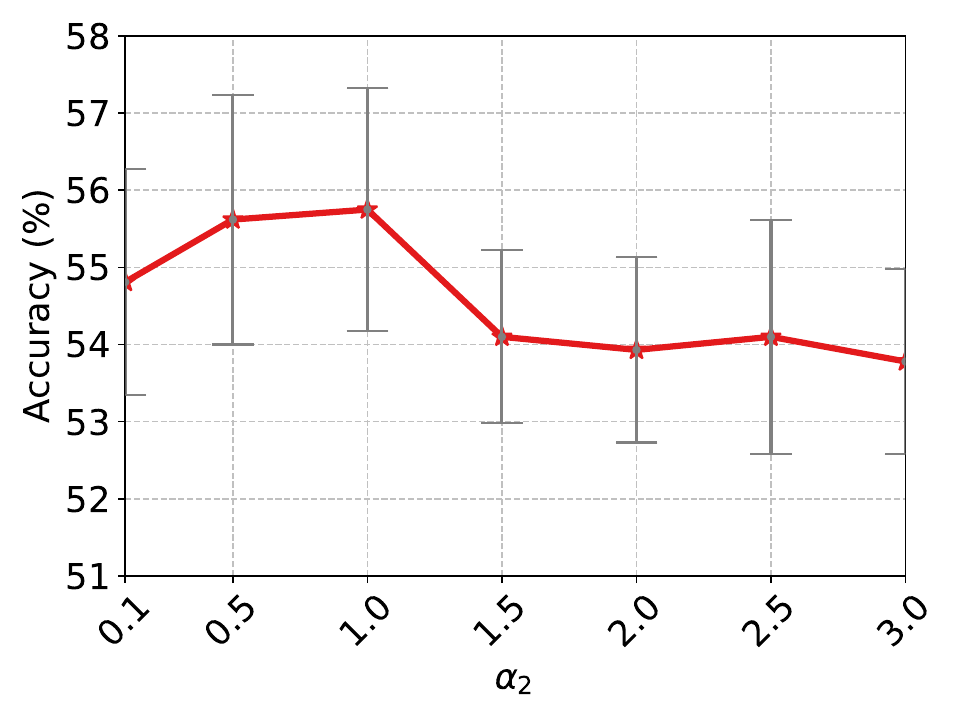}}
	\\
	\caption{Sensitivity analysis of L2D with respect to parameters $k$, $\alpha_1$, and $\alpha_2$ on the PACS dataset. The reported accuracy is the averaged over all three unseen target domains.}\label{fig:para_sens}
\end{figure*}

\subsection{Comparisons on Corrupted CIFAR-10}

\noindent\textbf{Experiment Setup}. We train all models on the training split of CIFAR-10 dataset (50,000 images) and evaluate them on the corrupted test split of CIFAR-10 (10,000 images). Following \cite{MADA}, we select WideResNet (16-4) \cite{wideresnet} as the backbone network with a batch size of 256. We optimize the model by using SGD with Nesterov momentum and weight decay rate of $0.0005$.  The learning rate is initialized to 0.2 which is gradually decreased by using the cosine annealing scheduler.

\begin{figure}[!t]
	\centering
	\subfloat[Noise]
	{\includegraphics[width=0.245\textwidth]{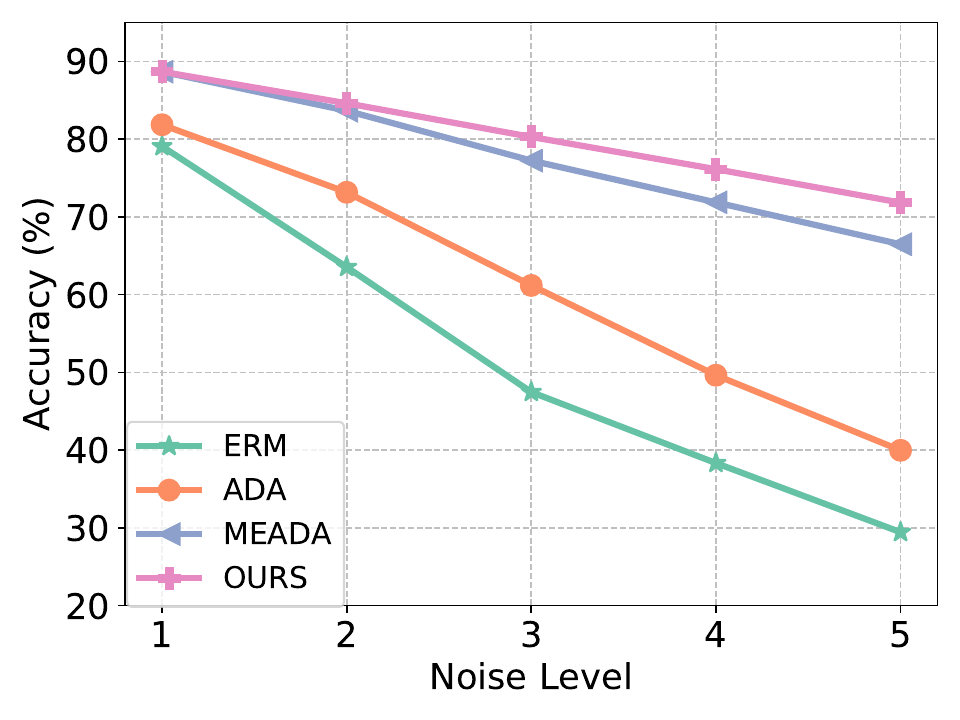}}
	\subfloat[Digital]
	{\includegraphics[width=0.245\textwidth]{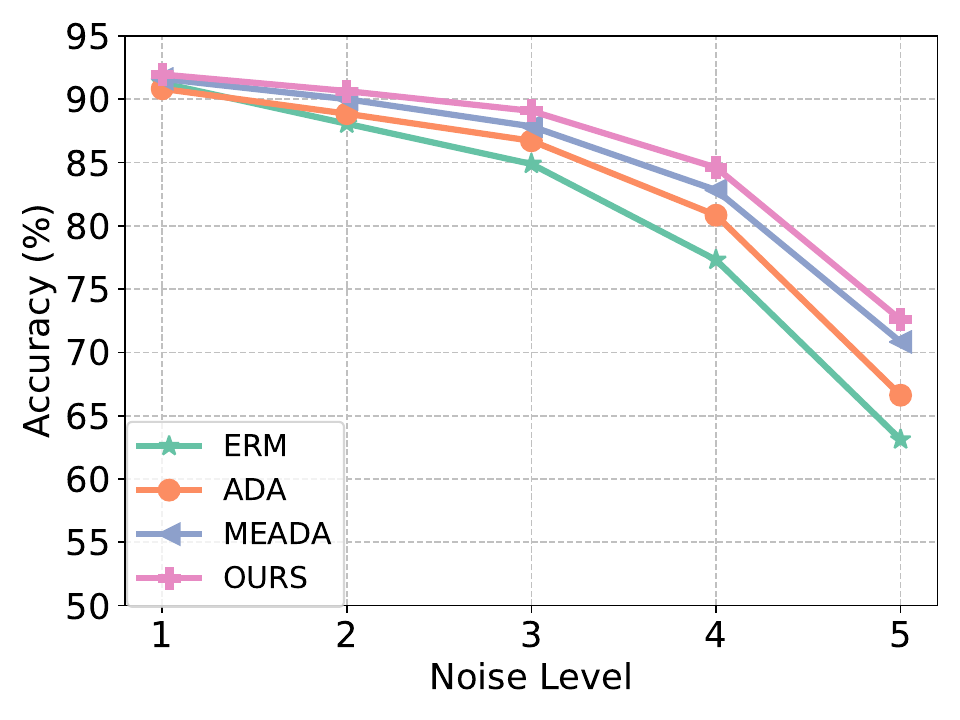}}
	\\
	\caption{Average classification accuracy (\%) of different methods under five levels of severity on \textit{Noise} and \textit{Digital} corruptions.}\label{fig: corruption}\vspace{-3ex}
\end{figure}

\noindent\textbf{Results}. We report the average accuracy of four categories under level-5 severity corruption in Table \ref{tab: cifar}. Our method achieves the highest average performance, surpassing the best baseline by approximately $2.6\%$. Notably, for noise corruptions, accuracy is significantly improved by approximately $13.2\%$. We also report the performance of methods against all five levels of Noise corruption and Digits corruption in Figure \ref{fig: corruption}. From the figure, we can see that the performance margin between our method and others are relatively small under severity level one, and it is gradually enlarged while the severity level goes up. This further validates that our model not only can achieve the highest average performance, but also is resilient towards the severe corruptions.   

\begin{table}[t]
\begin{center}
\caption{Single domain generalization accuracy (\%) on PACS. Models are trained on \textit{photo} and evaluated on the rest of the target domains (\textit{i.e., art painting, cartoon, and sketch}). The comparison is based on our implementation. Best performances are highlighted in bold.} 
\label{tab: single_pacs}
\resizebox{0.48\textwidth}{!}{
\begin{tabular}{l|ccc|c}
\toprule
& A & C & S & Avg.\\
\hline
ERM & 54.43 & 42.74 & 42.02 & 46.39 \\
JiGen & 54.98 & 42.62 & 40.62 & 46.07\\
RSC & 56.26 & 39.59 & 47.13 & 47.66\\
ADA & 58.72 & 45.58 & 48.26 & 50.85\\
ME-ADA & \textbf{58.96} & 44.09 & 49.96 & 51.00 \\
\hline
Ours w/o Style-comp. & 53.27 & 41.00 & 41.92 & 45.39 \\
Ours w/o Mod. & 58.48 & 48.96 & 53.20 & 53.54 \\
Ours w/o Min. MI & 56.49 & 48.08 & 56.32 & 53.63\\
Ours w/o Max. MI & 56.64 &  47.08 & 49.68 & 51.13\\
Ours (Full Model) & 56.26 & \textbf{51.04} & \textbf{58.42} & \textbf{55.24}\\
\bottomrule
\end{tabular}}
\end{center}\vspace{-3ex}
\end{table}

\subsection{Comparisons on PACS}
\noindent\textbf{Experiment Setup}. For the single domain generalization task, we consider a practical case, where we utilize a set of easy-to-collect realistic images (\textit{i.e.}, \textit{photo}) as the source domain, and evaluate models on the rest of diverse-styled domains (\textit{i.e.}, \textit{art painting}, \textit{cartoon}, and \textit{sketch}). AlexNet \cite{alexnet} is employed as the backbone which is pretrained on Imagenet and finetuned on the source domain. We also evaluate the effectiveness of our approach on PACS under the standard leave-one-domain-out protocol, where one domain is selected as the test domain and the rest are treated as the source domain. We employ pretrained Alexnet and ResNet-18 \cite{resnet} as the backbone networks for the leave-one-domain-out setting. Please refer to the supplementary material for more implementation details.


\noindent\textbf{Results}. From Table \ref{tab: single_pacs}, we can see that our method can achieve the best average classification accuracy compared to the baselines. Importantly, our method can achieve a relatively large performance margin over other methods on the \textit{sketch} domain, which has the largest domain shift from \textit{photo} due to its highly abstracted shapes. This result verifies that our method takes the advantage of diverse style images that are generated by the style-complement module.

To further validate the performance of our method, we conduct the leave-one-domain-out domain generalization task on PACS. We compare our approach with the two categories of recent state-of-the-art DG methods. The first category,including DSN \cite{dsn}, Fusion \cite{fusion}, MetaReg \cite{MetaReg}, Epi-FCR \cite{EPI-FCR}, MASF \cite{MASF} and DMG \cite{DMG}, requires domain identifications in the training phase. Methods in the second category are inline with a more realistic mixed latent domain setting \cite{mmld}, where domain identifiers are unavailable in the training phase. AGG \cite{EPI-FCR}, HEX \cite{HEX}, PAR \cite{PAR}, JiGen \cite{JIGSAW}, ADA \cite{ADA}, MEADA \cite{MEADA}, MMLD \cite{mmld} and our method belong to the latter category. We report the results with different backbone networks in Table \ref{tab:pacs}. Without leveraging the domain identifier, our method can still achieve the state-of-the-art performance on the leave-one-domain-out generalization task on PACS. In the training stage, the image augmentation module gradually enlarges the domain shift between the generated images and the source images.

\begin{table}
\begin{center}
\caption{Leave-one-domain-out classification accuracy(\%) on PACS. Best performances are highlighted in bold.} 
\label{tab:pacs}
\resizebox{0.48\textwidth}{!}{
\begin{tabular}{l|c|cccc|c}
\toprule
& D\_ID & P & A & C & S & Avg. \\
\midrule
\multicolumn{7}{l}{\textit{AlexNet}}\\
\midrule
DSN &\cmark & 83.30 & 61.10 & 66.50 & 58.60 & 67.40 \\
Fusion &\cmark & 90.20 & 64.10 & 66.80 & 60.10 & 70.30\\
MetaReg &\cmark & 87.40 & 63.50 & 69.50 & 59.10 & 69.90  \\
Epi-FCR &\cmark & 86.10 & 64.70 & 72.30 & 65.00 & 72.00\\
MASF &\cmark & 90.68 & 70.35 & 72.46 & 67.33 & 75.21\\
DMG &\cmark & 87.31 & 64.65 & 69.88 & \textbf{71.42} & 73.32\\
\hline
HEX &\xmark & 87.90 & 66.80 & 69.70 & 56.20 & 70.20\\
PAR &\xmark & 89.60 & 66.30 & 66.30 & 64.10 & 72.08\\
JiGen &\xmark &89.00& 67.63&71.71&65.18&73.38\\
ADA &\xmark & 85.10& 64.30& 69.80& 60.40 & 69.90\\
MEADA &\xmark & 88.60 & 67.10 & 69.90& 63.00 & 72.20\\
MMLD &\xmark & 88.98 & 69.27 & \textbf{72.83} & 66.44 & 74.38\\
\hline
Ours &\xmark &\textbf{90.96} &\textbf{71.19} & 72.18 &\textbf{67.68}&\textbf{75.50} \\
\midrule
\multicolumn{7}{l}{\textit{ResNet-18}}\\
\midrule
Epi-FCR &\cmark& 93.90 & \textbf{82.10} & 77.00 & 73.00 & 81.50 \\
MASF &\cmark& 94.99 & 80.29 & 77.17 & 71.68 & 81.03 \\
DMG &\cmark& 93.55 & 76.90 & \textbf{80.38} & 75.21 & 81.46 \\
\hline
Jigen &\xmark& 96.03 & 79.42 & 75.25 & 71.35 & 80.51 \\
ADA &\xmark& 95.61 & 78.32 &77.65 & 74.21 &81.44 \\
MEADA &\xmark& 95.57 & 78.61 & 78.65 & 75.59 & 82.10\\
MMLD &\xmark& \textbf{96.09} & 81.28 & 77.16 & 72.29 & 81.83\\
\hline
Ours &\xmark& 95.51 &\textbf{81.44} &\textbf{79.56} & \textbf{80.58}& \textbf{84.27} \\
\bottomrule

\end{tabular}}
\end{center}\vspace{-3ex}
\end{table}

\begin{figure*}[!t]
	\centering
	\subfloat[ERM]
	{\includegraphics[width=0.24\textwidth]{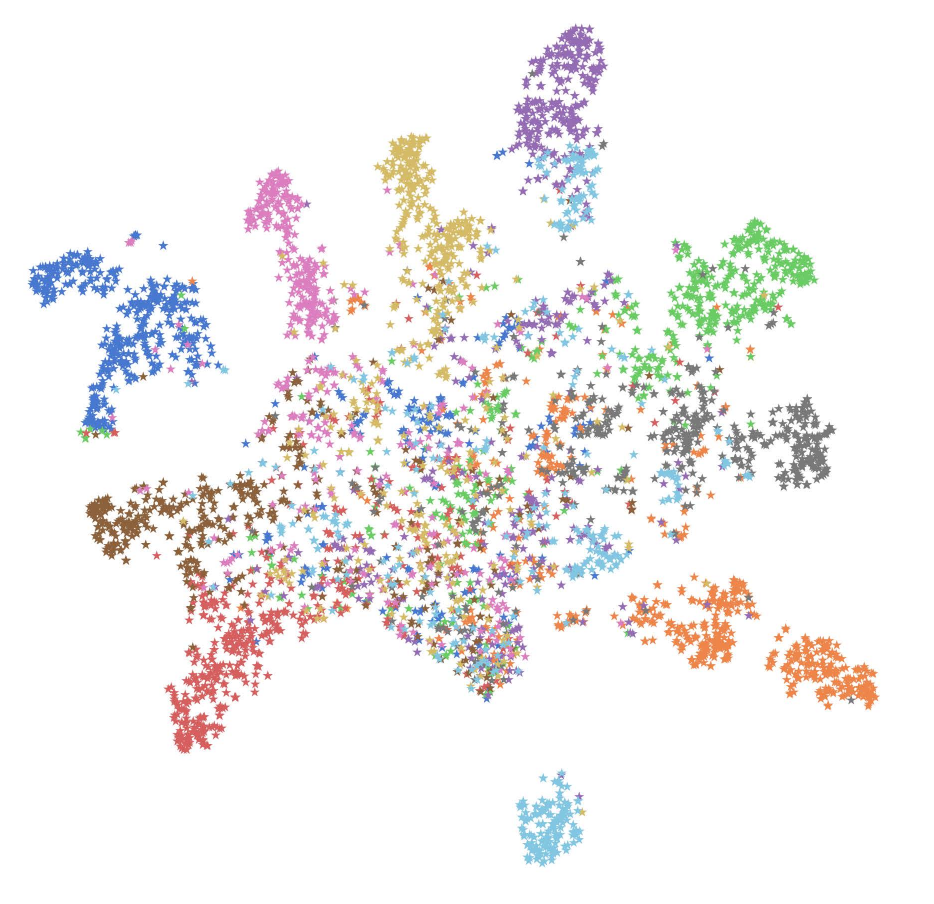}}
	\subfloat[ADA]
	{\includegraphics[width=0.24\textwidth]{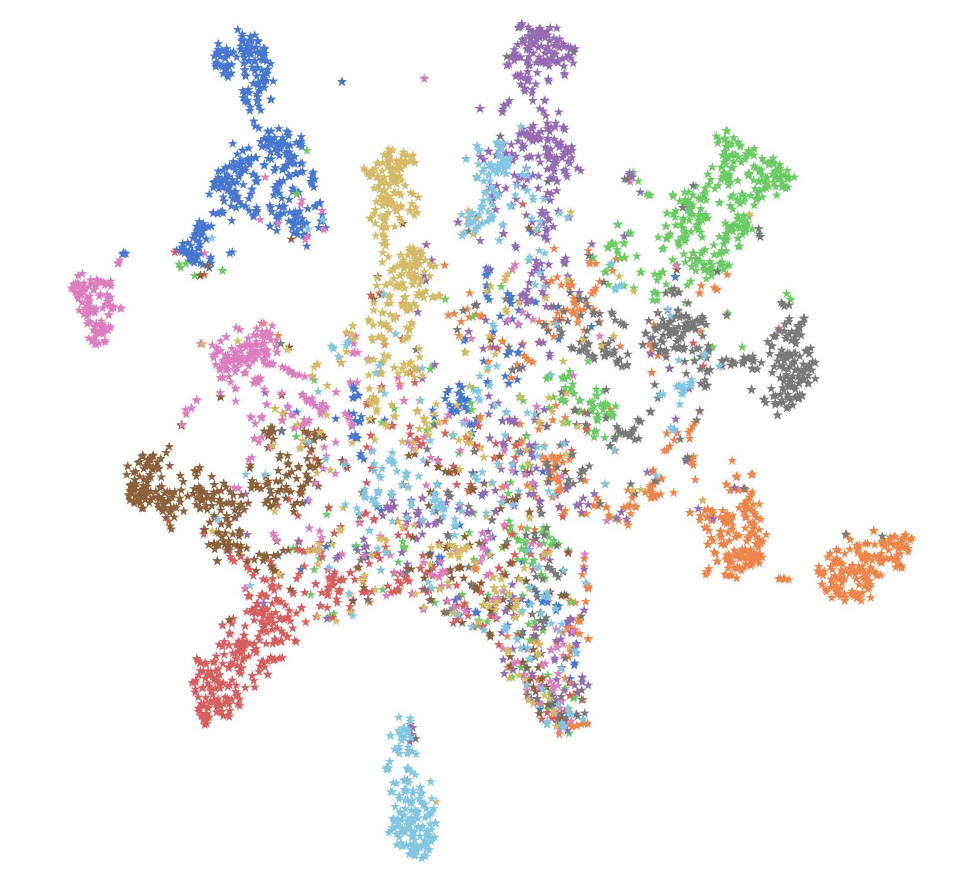}}
    \subfloat[ME-ADA]
	{\includegraphics[width=0.24\textwidth]{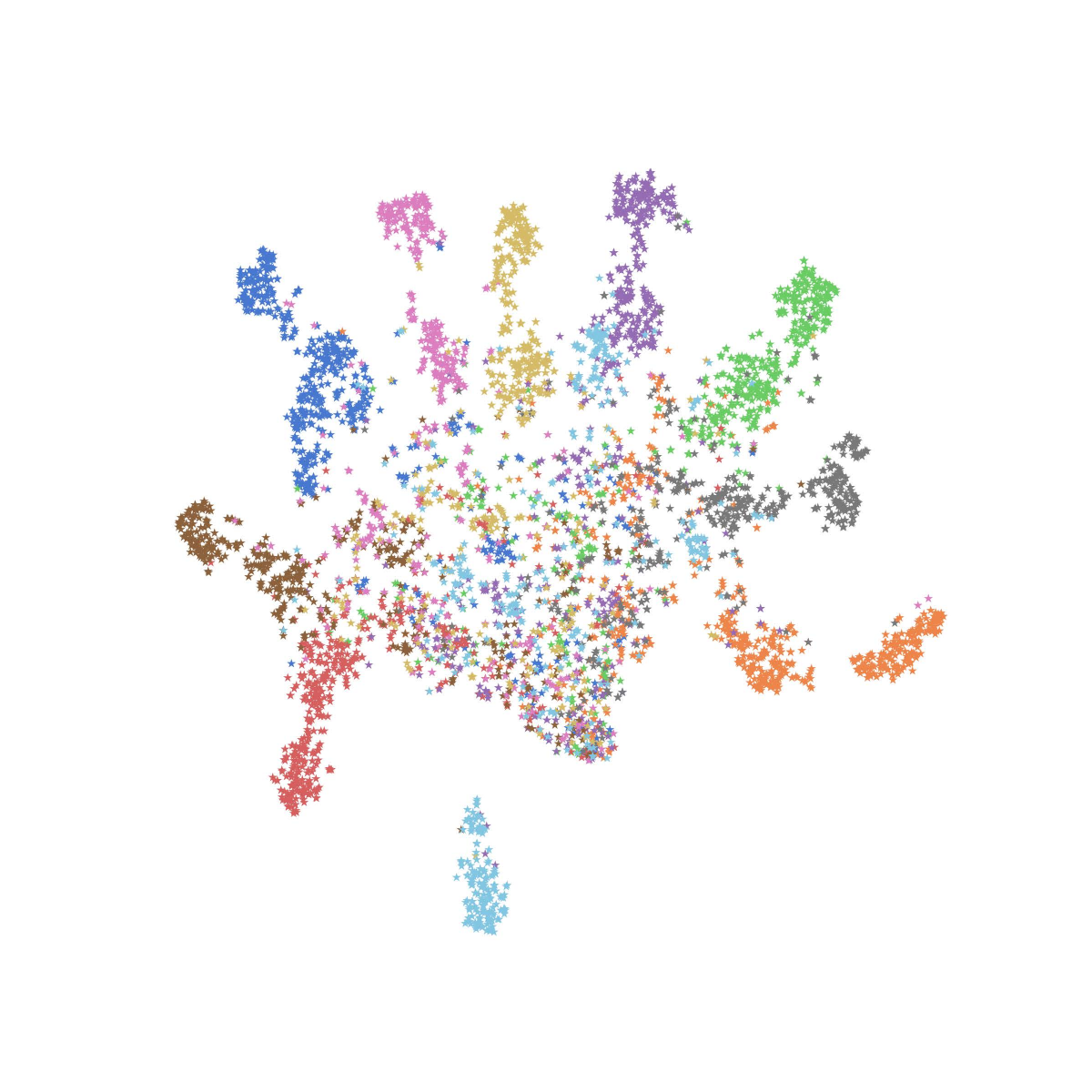}}
	\subfloat[OURS]
	{\includegraphics[width=0.24\textwidth]{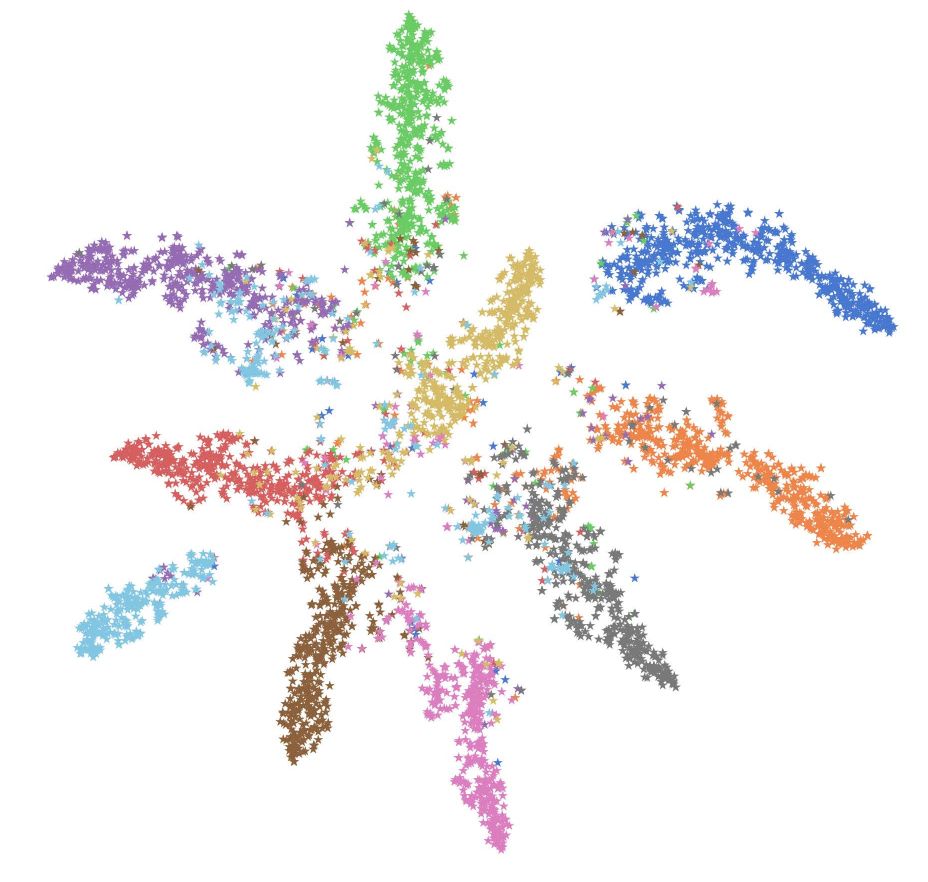}}
	\\
	\caption{The t-SNE visualizations of extracted unseen target feature distribution for different methods on digits Single-DG task. Features with the same semantic label are plotted in the same color.}\label{fig: tsne}
\end{figure*} 

\subsection{Empirical Analysis}
\noindent\textbf{Ablation Study}. 
We conduct the ablation study on PACS dataset to verify the effectiveness of each component in our framework. Table \ref{tab: single_pacs} reports the classification results of the 4 variants of our original framework. We report the baseline result without incorporating the style-complement module as \textit{w/o Style-comp.}. Without the generated images, the model is degraded to the backbone model with a variational approximation of $z$ in the embedding space. The large performance margin between \textit{w/o Style-comp.} and our full model demonstrates the importance of the style-complement module on improving the generalization ability of the model. \textit{w/o Mod.} shows the result after removing the style modification from the full model. The removal of the style modification triggers $1.35\%$ absolute performance decline. We infer this relates to the limited style diversity of the generated images. 

To provide an understanding of how mutual information affects the learning framework, in \textit{w/o Min. MI} and \textit{w/o Max. MI}, we remove the mutual information minimization and maximization process, respectively. We observe the accuracy of \textit{w/o Min. MI} is $~1.26\%$ lower than the full model which is similar to the results of \textit{w/o Mod.}. That means, without the mutual information constraint term, the style-complement module tends to generate images by following the source distribution, which limits their diversity. Meanwhile, without the mutual information maximization process, \textit{w/o Max. MI} a clear performance margin of $3.76\%$ is shown compared to the full model. This indicates that mapping diverse styled images from the same class closer in the embedding space is helpful for improving the generalization power of the model.

\noindent\textbf{Parameter Sensitivity}.
To validate the significance of the total number of transformations $k$, and weighting parameters $\alpha_1$, and $\alpha_2$ in the loss, we conduct sensitivity analysis on PACS dataset. In the experiments, we initially set $k=6$, $\alpha_1=1$, and $\alpha_2=1$. When we analyze the sensitivity to a specific parameter, we fix the values of the other two parameters. Figure \ref{fig:para_sens} shows the results of sensitivity analysis under the Single-DG setting. From Figure \ref{fig:para_sens}(a), we can see that the performance is gradually increasing as the style-complement module combines more transformations. Also, the average performance become relatively stable after $k=5$. The results also indicates that jointly considering multiple transformations increases the diversity of the generated images. Meanwhile, excessive transformations might produce extra noise, which hinders the further performance gain. As can be seen from Figure \ref{fig:para_sens} (b) and (c), our method surpasses the state-of-the-art performance in a wide range of $\alpha$ settings, and achieves the best classification accuracy with $\alpha_1=\alpha_2=1$. 

\noindent\textbf{t-SNE Visualizations}.
To further demonstrate the effectiveness of the proposed method, we use t-SNE \cite{tsne} to visualize the distribution of the unseen target features in the digits dataset (\textit{i.e.}, SVHN, SYN, USPS and MNIST-M). Specifically, we train different models on MNIST and randomly select 1000 samples from each unseen target domains to visualize. As shown in Fig. \ref{fig: tsne}, our method clearly achieve better class-wise separation than the baselines. Moreover, from the distribution of features extracted by ERM, ADA and ME-ADA, we observe that the features within the same class can have multiple sub-clusters. This indicates that it is hard for the methods to learn domain invariant representations and the large intra-class variations may hinder them obtaining a clear decision boundary on the targets. In contrast, it is clearly seen that our approach creates tighter clusters with a good class-wise mix compared to the baselines. This strongly supports the idea of diverse image generation and the maximization of mutual information among the samples from the same class.

\section{Conclusion}
This paper presented Learning-to-Diversify (L2D), a novel approach to address single domain generalization. Unlike previous domain generalization methods which exploit multiple source domains to learn domain invariant representations, the proposed approach designs a style-complement module to generate diverse out-of-domain images from single source domain. An iterative min-max mutual information (MI) optimization strategy is used to boost the generalization power of the model. The tractable MI upper bound is minimized to further enhance the diversity of the generated image, while MI among same category samples is maximized to obtain the style-invariant representations in the maximization step. Extensive experiments on three benchmark datasets demonstrate that the proposed method outperforms the state-of-the-art methods on both single domain generalization and the standard leave-one-domain-out domain generalization.

\noindent\textbf{Acknowledgement} This work was supported by Australian Research Council (ARC DP190102353, and CE200100025)

{\small
\balance{
\bibliographystyle{ieee_fullname}
\bibliography{main}

\begin{thebibliography}{10}\itemsep=-1pt

\bibitem{DIP}
Mahsa Baktashmotlagh, Mehrtash~Tafazzoli Harandi, Brian~C. Lovell, and Mathieu
  Salzmann.
\newblock Unsupervised domain adaptation by domain invariant projection.
\newblock In {\em ICCV}, 2013.

\bibitem{MetaReg}
Yogesh Balaji, Swami Sankaranarayanan, and Rama Chellappa.
\newblock Metareg: Towards domain generalization using meta-regularization.
\newblock In {\em NeurIPS}, 2018.

\bibitem{discrepancybendavid}
Shai Ben{-}David, John Blitzer, Koby Crammer, Alex Kulesza, Fernando Pereira,
  and Jennifer~Wortman Vaughan.
\newblock A theory of learning from different domains.
\newblock {\em Mach. Learn.}, 2010.

\bibitem{dsn}
Konstantinos Bousmalis, George Trigeorgis, Nathan Silberman, Dilip Krishnan,
  and Dumitru Erhan.
\newblock Domain separation networks.
\newblock In {\em NeurIPS}, 2016.

\bibitem{JIGSAW}
Fabio~Maria Carlucci, Antonio D'Innocente, Silvia Bucci, Barbara Caputo, and
  Tatiana Tommasi.
\newblock Domain generalization by solving jigsaw puzzles.
\newblock In {\em CVPR}, 2019.

\bibitem{DMG}
Prithvijit Chattopadhyay, Yogesh Balaji, and Judy Hoffman.
\newblock Learning to balance specificity and invariance for in and out of
  domain generalization.
\newblock In {\em ECCV}, 2020.

\bibitem{CLUB}
Pengyu Cheng, Weituo Hao, Shuyang Dai, Jiachang Liu, Zhe Gan, and Lawrence
  Carin.
\newblock {CLUB:} {A} contrastive log-ratio upper bound of mutual information.
\newblock In {\em ICML}, 2020.

\bibitem{discrepancy2}
Gabriela Csurka.
\newblock A comprehensive survey on domain adaptation for visual applications.
\newblock In {\em Domain Adaptation in Computer Vision Applications}. 2017.

\bibitem{USPS}
John~S. Denker, W.~R. Gardner, Hans~Peter Graf, Donnie Henderson, Richard~E.
  Howard, Wayne~E. Hubbard, Lawrence~D. Jackel, Henry~S. Baird, and Isabelle
  Guyon.
\newblock Neural network recognizer for hand-written zip code digits.
\newblock In {\em NeurIPS}, 1988.

\bibitem{MASF}
Qi Dou, Daniel~Coelho de Castro, Konstantinos Kamnitsas, and Ben Glocker.
\newblock Domain generalization via model-agnostic learning of semantic
  features.
\newblock In {\em NeurIPS}, 2019.

\bibitem{MetaVIB}
Ying{-}Jun Du, Jun Xu, Huan Xiong, Qiang Qiu, Xiantong Zhen, Cees G.~M. Snoek,
  and Ling Shao.
\newblock Learning to learn with variational information bottleneck for domain
  generalization.
\newblock In {\em ECCV}, 2020.

\bibitem{SYN}
Yaroslav Ganin and Victor~S. Lempitsky.
\newblock Unsupervised domain adaptation by backpropagation.
\newblock In {\em ICML}, 2015.

\bibitem{DANN}
Yaroslav Ganin, Evgeniya Ustinova, Hana Ajakan, Pascal Germain, Hugo
  Larochelle, Fran{\c{c}}ois Laviolette, Mario Marchand, and Victor~S.
  Lempitsky.
\newblock Domain-adversarial training of neural networks.
\newblock {\em J. Mach. Learn. Res.}, 2016.

\bibitem{MTAE}
Muhammad Ghifary, W.~Bastiaan Kleijn, Mengjie Zhang, and David Balduzzi.
\newblock Domain generalization for object recognition with multi-task
  autoencoders.
\newblock In {\em ICCV}, 2015.

\bibitem{resnet}
Kaiming He, Xiangyu Zhang, Shaoqing Ren, and Jian Sun.
\newblock Deep residual learning for image recognition.
\newblock In {\em CVPR}, 2016.

\bibitem{cifar-c}
Dan Hendrycks and Thomas Dietterich.
\newblock Benchmarking neural network robustness to common corruptions and
  perturbations.
\newblock {\em ICLR}, 2019.

\bibitem{RSC}
Zeyi Huang, Haohan Wang, Eric~P. Xing, and Dong Huang.
\newblock Self-challenging improves cross-domain generalization.
\newblock In {\em ECCV}, 2020.

\bibitem{semiDA}
Hal~Daum{\'{e}} III, Abhishek Kumar, and Avishek Saha.
\newblock Co-regularization based semi-supervised domain adaptation.
\newblock In {\em NeurIPS}, 2010.

\bibitem{SUPCON}
Prannay Khosla, Piotr Teterwak, Chen Wang, Aaron Sarna, Yonglong Tian, Phillip
  Isola, Aaron Maschinot, Ce Liu, and Dilip Krishnan.
\newblock Supervised contrastive learning.
\newblock In {\em NeurIPS}, 2020.

\bibitem{cifar}
Alex Krizhevsky, Geoffrey Hinton, et~al.
\newblock Learning multiple layers of features from tiny images.
\newblock 2009.

\bibitem{alexnet}
Alex Krizhevsky, Ilya Sutskever, and Geoffrey~E. Hinton.
\newblock Imagenet classification with deep convolutional neural networks.
\newblock In {\em NeurIPS}, 2012.

\bibitem{MNIST}
Yann LeCun, L{\'e}on Bottou, Yoshua Bengio, and Patrick Haffner.
\newblock Gradient-based learning applied to document recognition.
\newblock {\em Proceedings of the IEEE}, 1998.

\bibitem{pacs}
Da Li, Yongxin Yang, Yi{-}Zhe Song, and Timothy~M. Hospedales.
\newblock Deeper, broader and artier domain generalization.
\newblock In {\em ICCV}, 2017.

\bibitem{MLDG}
Da Li, Yongxin Yang, Yi{-}Zhe Song, and Timothy~M. Hospedales.
\newblock Learning to generalize: Meta-learning for domain generalization.
\newblock In {\em AAAI}, 2018.

\bibitem{EPI-FCR}
Da Li, Jianshu Zhang, Yongxin Yang, Cong Liu, Yi{-}Zhe Song, and Timothy~M.
  Hospedales.
\newblock Episodic training for domain generalization.
\newblock In {\em ICCV}, 2019.

\bibitem{DAN}
Mingsheng Long, Yue Cao, Jianmin Wang, and Michael~I. Jordan.
\newblock Learning transferable features with deep adaptation networks.
\newblock In {\em ICML}, 2015.

\bibitem{CDAN}
Mingsheng Long, Zhangjie Cao, Jianmin Wang, and Michael~I. Jordan.
\newblock Conditional adversarial domain adaptation.
\newblock In {\em NeurIPS}, 2018.

\bibitem{JAN}
Mingsheng Long, Han Zhu, Jianmin Wang, and Michael~I. Jordan.
\newblock Deep transfer learning with joint adaptation networks.
\newblock In {\em ICML}, 2017.

\bibitem{PGL}
Yadan Luo, Zijian Wang, Zi Huang, and Mahsa Baktashmotlagh.
\newblock Progressive graph learning for open-set domain adaptation.
\newblock In {\em ICML}, 2020.

\bibitem{fusion}
Massimiliano Mancini, Samuel~Rota Bul{\`{o}}, Barbara Caputo, and Elisa Ricci.
\newblock Best sources forward: Domain generalization through source-specific
  nets.
\newblock In {\em ICIP}, 2018.

\bibitem{mmld}
Toshihiko Matsuura and Tatsuya Harada.
\newblock Domain generalization using a mixture of multiple latent domains.
\newblock In {\em AAAI}, 2020.

\bibitem{fewshotDA}
Saeid Motiian, Quinn Jones, Seyed~Mehdi Iranmanesh, and Gianfranco Doretto.
\newblock Few-shot adversarial domain adaptation.
\newblock In {\em NeurIPS}, 2017.

\bibitem{CCSA}
Saeid Motiian, Marco Piccirilli, Donald~A. Adjeroh, and Gianfranco Doretto.
\newblock Unified deep supervised domain adaptation and generalization.
\newblock In {\em ICCV}, 2017.

\bibitem{DICA}
Krikamol Muandet, David Balduzzi, and Bernhard Sch{\"{o}}lkopf.
\newblock Domain generalization via invariant feature representation.
\newblock In {\em ICML}, 2013.

\bibitem{SVHN}
Yuval Netzer, Tao Wang, Adam Coates, Alessandro Bissacco, Bo Wu, and Andrew~Y
  Ng.
\newblock Reading digits in natural images with unsupervised feature learning.
\newblock 2011.

\bibitem{discrepancy1}
Sinno~Jialin Pan and Qiang Yang.
\newblock A survey on transfer learning.
\newblock {\em {IEEE} Trans. Knowl. Data Eng.}, 2010.

\bibitem{MADA}
Fengchun Qiao, Long Zhao, and Xi Peng.
\newblock Learning to learn single domain generalization.
\newblock In {\em CVPR}, 2020.

\bibitem{semiDA2}
Kuniaki Saito, Donghyun Kim, Stan Sclaroff, Trevor Darrell, and Kate Saenko.
\newblock Semi-supervised domain adaptation via minimax entropy.
\newblock In {\em ICCV}, 2019.

\bibitem{CROSSGRAD}
Shiv Shankar, Vihari Piratla, Soumen Chakrabarti, Siddhartha Chaudhuri, Preethi
  Jyothi, and Sunita Sarawagi.
\newblock Generalizing across domains via cross-gradient training.
\newblock In {\em ICLR}, 2018.

\bibitem{infonce}
A{\"{a}}ron van~den Oord, Yazhe Li, and Oriol Vinyals.
\newblock Representation learning with contrastive predictive coding.
\newblock {\em CoRR}, abs/1807.03748, 2018.

\bibitem{tsne}
Laurens Van~der Maaten and Geoffrey Hinton.
\newblock Visualizing data using t-sne.
\newblock {\em J. Mach. Learn. Res.}, 2008.

\bibitem{ADA}
Riccardo Volpi, Hongseok Namkoong, Ozan Sener, John~C. Duchi, Vittorio Murino,
  and Silvio Savarese.
\newblock Generalizing to unseen domains via adversarial data augmentation.
\newblock In {\em NeurIPS}, 2018.

\bibitem{PAR}
Haohan Wang, Songwei Ge, Zachary~C. Lipton, and Eric~P. Xing.
\newblock Learning robust global representations by penalizing local predictive
  power.
\newblock In {\em NeurIPS}, 2019.

\bibitem{HEX}
Haohan Wang, Zexue He, Zachary~C. Lipton, and Eric~P. Xing.
\newblock Learning robust representations by projecting superficial statistics
  out.
\newblock In {\em ICLR}, 2019.

\bibitem{DAsurvey}
Mei Wang and Weihong Deng.
\newblock Deep visual domain adaptation: {A} survey.
\newblock {\em Neurocomputing}, 2018.

\bibitem{PMGN}
Zijian Wang, Yadan Luo, Zi Huang, and Mahsa Baktashmotlagh.
\newblock Prototype-matching graph network for heterogeneous domain adaptation.
\newblock In {\em MM}, 2020.

\bibitem{dsne}
Xiang Xu, Xiong Zhou, Ragav Venkatesan, Gurumurthy Swaminathan, and Orchid
  Majumder.
\newblock d-sne: Domain adaptation using stochastic neighborhood embedding.
\newblock In {\em CVPR}, 2019.

\bibitem{randconv}
Zhenlin Xu, Deyi Liu, Junlin Yang, Colin Raffel, and Marc Niethammer.
\newblock Robust and generalizable visual representation learning via random
  convolutions.
\newblock In {\em ICLR}, 2021.

\bibitem{wideresnet}
Sergey Zagoruyko and Nikos Komodakis.
\newblock Wide residual networks.
\newblock In {\em BMVC}, 2016.

\bibitem{MEADA}
Long Zhao, Ting Liu, Xi Peng, and Dimitris~N. Metaxas.
\newblock Maximum-entropy adversarial data augmentation for improved
  generalization and robustness.
\newblock In {\em NeurIPS}, 2020.

\bibitem{L2A}
Kaiyang Zhou, Yongxin Yang, Timothy~M. Hospedales, and Tao Xiang.
\newblock Learning to generate novel domains for domain generalization.
\newblock In {\em ECCV}, 2020.

\end{thebibliography}
}
}

\end{document}